\def\year{2017}\relax
\documentclass[letterpaper]{article}
\usepackage{aaai17}
\usepackage{times}
\usepackage{helvet}
\usepackage{courier}

\usepackage{times}
\usepackage{latexsym}
\usepackage{graphicx, graphics}
\usepackage{algorithmic}
\usepackage{tabularx}
\usepackage{multirow}
\usepackage{subfigure}
\usepackage{amsmath}
\usepackage{amsfonts}

\title{Word Embedding Based Correlation Model for Question/Answer Matching}
\author{
	Yikang Shen$^1$, Wenge Rong$^2$, Nan Jiang$^2$, Baolin Peng$^3$, Jie Tang$^4$, Zhang Xiong$^2$\\
	$^1$Sino-French Engineer School, Beihang University, China\\
	$^2$School of Computer Science and Engineering, Beihang University, China\\
	$^3$Department of System Engineering and Engineering Management, The Chinese University of Hong Kong, China\\
	$^4$Department of Computer Science and Technology, Tsinghua University, China\\
	\{yikang.shen, w.rong, nanjiang, xiongz\}@buaa.edu.cn, blpeng@se.cuhk.edu.hk, jietang@tsinghua.edu.cn
}

\begin{document}

\maketitle

\begin{abstract}
The large scale of Q\&A archives accumulated in community based question answering (CQA) servivces are important information and knowledge resource on the web. Question and answer matching task has been attached much importance to for its ability to reuse knowledge stored in these systems: it can be useful in enhancing user experience with recurrent questions. In this paper, a Word Embedding based Correlation (WEC) model is proposed by integrating advantages of both the translation model and word embedding. Given a random pair of words, WEC can score their co-occurrence probability in Q\&A pairs, while it can also leverage the continuity and smoothness of continuous space word representation to deal with new pairs of words that are rare in the training parallel text. An experimental study on Yahoo! Answers dataset and Baidu Zhidao dataset shows this new method's promising potential.
\end{abstract}

\section{Introduction}

Community Question Answering (CQA) services are websites that enable users to share knowledge by asking and answering different kinds of questions. Over the last decade, websites, such as Yahoo! Answers, Baidu Zhidao, Quora, and Zhihu, have accumulated large scale question and answer (Q\&A) archives, which are usually organised as a question with a list of candidate answers and associated with metadata including user tagged subject categories, answer popularity votes, and selected correct answer \cite{ZhouHe-156}. This user-generated content is an important information repository on the web and makes Q\&A archives invaluable resources for various tasks such as question-answering \cite{JeonCroft-67,XueJeon-87,alessandromoschitti2015semeval,alessandromoschittisemeval} and knowledge mining \cite{AdamicZhang-154}.

To make better use of information stored in CQA systems, a fundamental task is to properly matching potential candidate answers to the question, since many questions recur enough to allow for at least a few new questions to be answered by past materials \cite{DBLP:conf/www/ShtokDMS12}. There are several challenges for this task among which the lexical gap or lexical chasm between the question and candidate answers is a difficult one \cite{BergerCaruana-162}. Lexical gap describes the distance between dissimilar but potentiality related words in questions and answers. For example, given the question ``What is the fastest car in the world?'', a good answer might be ``The Jaguar XJ220 is the dearest, fastest and most sought after car on the planet.'' This Q\&A pair share no more than 4 words in common, including ``the'' and ``is'', but they are strongly associated by synonyms, hyponyms, or other weaker semantic associations \cite{YihHe-150}. Due to the heterogeneity of question and answer, the lexical gap is more significant in Q\&A matching task than other paraphrases detection task or information retrieval task.

A possible approach for the lexical gap problem is to employ translation model, which will leverage the Q\&A pairs to learn the semantically related words \cite{JeonCroft-67,XueJeon-87,ZhouLyu-50,guzman-marquez-nakov:2016:P16-2}. The basic assumption is that Q\&A pairs are ``parallel text'' and relationship between words (or phrases) can be established through word-to-word (or phrase-to-phrase) translation probabilities by representing words in a discrete space. In spite of its wide use in many natural language processing tasks, discrete space representation has two majors disadvantages: 1) the curse of dimensionality \cite{BengioDucharme-163}, for a natural language with a vocabulary $V$ of size $N$, we need to learn at most $N^{N}$ word-to-word translation probabilities; 2) the generalisation structure is not obvious: it is difficult to estimate the probability of exact word if they are rare in the training parallel text \cite{ZouSocher-158}.

An alternative method is to use a semantic-based model. Some work proposed to learn the latent topics aligned across the question-answer pairs to bridge the lexical gap, with the assumption that a question and its answer should share similar topic distribution \cite{CaiZhou-160,JiXu-161}. Recently, inspired by the success of word embedding, some papers propose to leverage the advantage of the vector representation of words to overcome the lexical gap \cite{ShenRong-155,ZhouHe-156} by using similarity of word vector to represent the word-to-word relation. In other words, this method calculates Q\&A matching probability based on semantic similarities between words. Because local smoothness properties of continuous space word representations, generalisation can be obtain more easily \cite{BengioDucharme-163}. However question and answers are heterogeneous in many aspects, semantic similarities can be weak between questions and answers \cite{ZhouHe-156}.

Inspired by the pros and cons of the translation model and semantic model, in this paper we propose a Word Embedding Correlation (WEC) model, which integrates the advantages of both the translation model \cite{JeonCroft-67,XueJeon-87} and word embedding \cite{BengioDucharme-163,MikolovChen-63,pennington2014glove}. In this model, a word-level correlations function $C(q_i,a_j)$ is designed to capture the word-to-word relation. Similar to traditional translation probability, this function calculates words co-occurrence probability in parallel text (Q\&A pairs). Instead of using word's discrete representation and maintaining a big and sparse translation probability matrix, we map input words $q_i$ and $a_j$ into vectors and use a low dimension dense translation matrix $\mathbf{M}$ to capture the co-occurrence relationship of words. If co-occurrences of exact words are rare in the training parallel text, $C(q_i,a_j)$ can also estimate their correlations strength because of the local smoothness properties of continuous space word representations \cite{BengioDucharme-163}. Based on the word-level correlations function, we propose a sentence-level correlations functions $C(q,a)$ to calculate the relevance between question and answer. This sentence-level correlation function also makes it possible to learn the translation matrix $\mathbf{M}$ directly from parallel corpus. Furthermore, we combine our model with convolution neural network (CNN) \cite{LecunBottou-98,ShenRong-155} to integrate both lexical and syntactical information stored in Q\&A to estimate the matching probability. Experimental study on Yahoo! Answers dataset and Baidu Zhidao dataset has shown WEC model's potential.

The proposed model will be illustrated in detail in Section 2. Section 3 will elaborate on the experimental study. Section 4 will present related work in solving the CQA matching problem and Section 5 concludes the paper and highlights possible future research directions.

\section{Methodology}

\textbf{Problem Definition} \textit{Given a question $q = q_1...q_n$, where $q_i$ is the i-th word in the question, and a set of candidate answers $\mathcal{A} = \{a^1, a^2, ..., a^n\}$, where $a^j = a^j_1...a^j_m$ and $a^j_k$ is the k-th word in j-th candidate answer, the goal is to identify the most relevant answer $a^{best}$.} 

In order to solve this problem, we calculate the matching probability between $q$ and each answer $a^i$, and then rank candidate answers by their matching probabilities, which are calculated through three steps: 1) words in questions and answers are represented by vectors in a continuous space; 2) word-to-word correlation score is calculated by using a word-level correlation function; 3) Q\&A matching probability is obtained by employing a phrase-level correlation function. Furthermore, we also propose to incorporate the proposed WEC model with convolution neural network (CNN) to achieve a better matching precision.

\subsection{Word Embedding}
In order to properly represent words in a continuous space, the idea of a neural language model \cite{BengioDucharme-163} is employed to enable jointly learn embedding of words into an $n$-dimensional vector space and to use these vectors to predict how likely a word is given its context. Skip-gram model \cite{MikolovChen-63} is a widely used approach to compute such an embedding. When skip-gram networks are optimised via gradient ascent, the derivatives modify the word embedding matrix $L \in R^{(n \times |V|)}$, where $|V|$ is the size of the vocabulary. The word vectors inside the embedding matrix capture distributional syntactic and semantic information via the word co-occurrence statistics \cite{BengioDucharme-163,MikolovChen-63}. Once this matrix is learned on an unlabelled corpus, it can be used for subsequent tasks by using each word's vector $v_w$ (a column in $L$) to represent that word.

\subsection{Word Embedding based Correlation (WEC) Model}
\subsubsection{Word-level Correlation Function}

In this paper we try to discover a correlation scoring function that uses word embedding as input and can also model the co-occurrence of words at the same time. In order to achieve this goal, we use a translation matrix $\mathbf{M}$ to transform words in the answer into words in the question. Given a pair of words $(q_i,a_j)$, their WEC scoring function is defined as:
\begin{equation} \label{word-level function}
C(q_i,a_j) = \cos<v_{q_i}, \mathbf{M} v_{a_j}> = \frac{v_{q_i}^\mathrm{T}}{||v_{q_i}||} \frac{\mathbf{M} v_{a_j}}{||\mathbf{M} v_{a_j}||}
\end{equation}
where $v_{q_i}$ and $v_{a_j}$ represent $q_i$ and $a_j$'s $d$-dimensional word embedding vectors; $||\cdot||$ is Euclidean norm; correlations matrix $\mathbf{M} \in \mathbb{R} ^ {d \times d}$. $\mathbf{M}$ is called translation matrix, because it maps word in the answer into a possible correlated word in the question. Then the cosine function will be used to capture the semantic similarity between origin words in the question and the mapped words. 
Previous cosine similarity is a special case of WEC Scoring Function when $\mathbf{M}$ is set to identity matrix. Meanwhile, $C(q_i,a_j)$ does not necessarily equal to $C(a_j,q_i)$, because the probability of $q_i$ existing in question and $a_j$ existing in answer may not equal to the probability of $a_j$ existing in question and $q_i$ existing in answer.

\subsubsection{Sentence-level Correlation Function}
Based on the word-level correlation function, we further propose the sentence-level correlation function, which integrates word-to-word correlation scores into the Q\&A pair correlation score.
For a Q\&A pair $(q,a)$, their correlation score is defined as:
\begin{equation} \label{sentence-level function}
C(q,a) = \frac{1}{|a|} \sum_j \max_i C(q_i,a_j)
\end{equation}
where $|a|$ represents the length of answer $a$, $C(q_i,a_j)$ is the correlations score of the $i$-th word in question and the $j$-th word in the answer. The max-operator choose one most related word in question for each word in answer. Sentence level correlation score is calculated by averaging selected word-level scores. According to our experiments, this max-average function perform better than simply average all word-level correlation score, and maximizing more efficiently.

\subsection{WEC + Convolution Neural Networks (CNN)}
WEC is based on the bag-of-word schema, which puts the syntactical information aside, e.g., the word sequence information. As such in worst cases, two phrases may have same bag-of-words representation, their real meaning could be completely opposite \cite{SocherPennington-93}. 

\begin{figure}[!h]
  \centering
  \includegraphics[width=1\columnwidth]{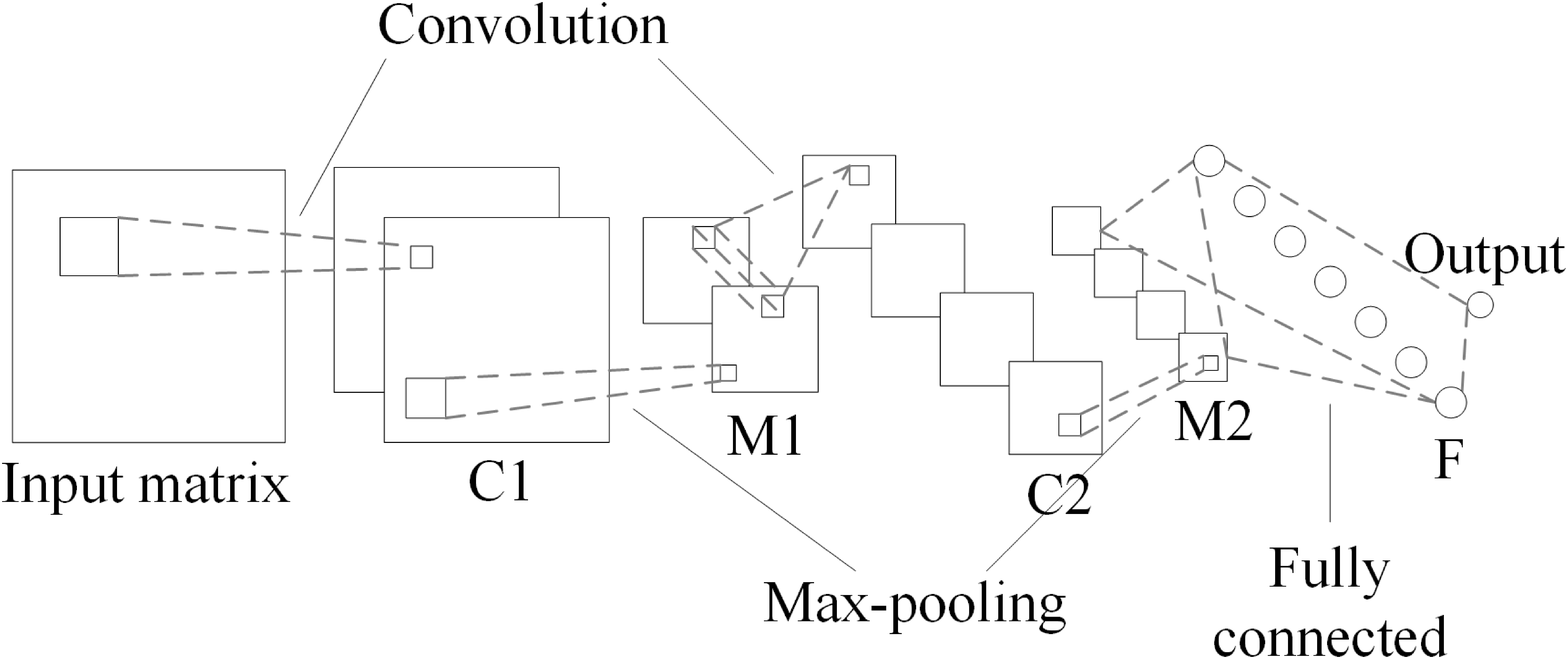}
  \caption{Architecture of CNN. It has two convolution layers, C1 and C2, each convolution layer is followed by a max-pooling layer, P1 and P2, and fully connected layer F. The input matrix ($\mathbf{S}$ or $\mathbf{C}$) is an $n_f \times m_f$ fixed size matrix.}
  \label{picture3}
\end{figure}

To overcome this limitation, several approaches have been proposed and one possible solution is to use the convolution neural network (CNN) model \cite{he2015multi,mou2016convolutional}. \cite{DBLP:conf/acl/KalchbrennerGB14} proposed that the convolutional and dynamic pooling layer in CNN can relate phrases far apart in the input sentence. In \cite{ShenRong-155}, $S$+CNN model is proposed for Q\&A matching to integrate both syntactical and lexical information to estimate the matching probability. In their model, the input Q\&A pair is transformed into a similarity matrix $\mathbf{S}$, generated through function:
\begin{equation}
\mathbf{S}_{ij}=cos(q_{i \bmod |q|},a_{j \bmod |a|})
\end{equation}
where $|q|$ and $|a|$ are the respective lengths of question and answer, $\mathbf{S}$ is a $n_f \times m_f$ fix size matrix, and $n_f$ and $m_f$ are the number of rows and columns respectively. Thus, the maximum length of questions and answers should be limited to be no longer than $n_f$ and $m_f$.  Then the similarity matrix is used as an input of a CNN instead of an image in \cite{LecunBottou-98}, the output of the CNN is the matching score of the Q\&A pair. Fig. \ref{picture3} shows the architecture of the employed CNN.

Instead of word embedding cosine similarity, the word-level correlation scores in WEC can be used in the formation of the input matrix of CNN. Similar to the $S$-matrix, we propose a correlations matrix $\mathbf{C}$, generated through function:
\begin{equation}
\mathbf{C}_{ij}=C(q_{i \bmod |q|},a_{j \bmod |a|})
\end{equation}
where $\mathbf{C}$ is an $n_f \times m_f$ fixed size matrix, and used as the input matrix of CNN. In this way, we obtain a new combination model, called the WEC+CNN model.

The complete training process comprises two supervised pre-training steps and a supervised fine-tuning step. In the first supervised pre-training step, we maximize the margin of the output of WEC function to pre-train $M$. In the second supervised pre-training step, we fix $M$, then maximize the margin of the output of CNN to train the CNN part. In the fine-tuning step, we maximize the margin of the output of CNN to fine-tune all parameters in WEC and CNN.

\section{Experimental Study}
\subsection{Dataset}
To evaluate the proposed WEC model, two datasets Yahoo! Answer and Baidu Zhidao are employed in this research. The dataset from Yahoo! Answers is available in Yahoo! Webscope\footnote{http://research.yahoo.com/Academic\_Relations}, including a large number of questions and their corresponding answers. In addition, the corpus contains a small amount of meta data, such as, which answer was selected as the best answer, and the category and sub-category assigned to this question \cite{SurdeanuCiaramita-77}. To validate the proposed WEC model, we generate three different sub-datasets. As shown in Table \ref{tb4}, the first subset contains 53,261 questions which are categorized as being about travel. The second subset contains 57,576 questions that are categorized as being about relationships. The third subset contains 66,129 questions that are categorized as being about finance. Because the CNN model needs to limit the maximum length of questions and answers, all selected questions are no longer than 50 words, selected answers a no longer than 100 words. Thus, for the Yahoo! Answer dataset, $n_f=50$ and $m_f=100$. We also limit the minimum length of an answer to 5 words, to avoid answers like ``Yes", ``It doesn't work" or simply a URL. More than half of the questions and answers in Yahoo! Answers satisfy these limitations.

The Baidu Zhidao dataset is provided in \cite{ShenRong-155}, and contains 99,909 questions and their best answers. Following their settings, 4 different datasets are generated, each contains questions from a single category: 'Computers \& Internet', 'Education \& Science', 'Games', and 'Entertainment \& Recreation'. each dataset contains 90,000 training triples, the random category dataset contains 10,000 test questions, other datasets contain 1,000 test questions each. In this dataset $n_f=30$ and $m_f=50$.

\begin{table*}[!ht]
	\center
	\caption{Compositions of Yahoo! Answer dataset. We randomly split questions from each category into training sets, validation sets and test sets by 4:1:1. For each question $q$ in training (or validation) sets, 10 different $(q, a^+, a^-)$ triples are generated, where $a^+$ is $q$'s best answer, $y^-$ is randomly selected answer from other question in the same category. Same $q$ and $a^+$ appear in the 10 triples with different $a^-$.}
	\begin{tabularx}{170mm}{|X|c|c|c|c|c|c|}
		\hline
		Category 		& Question\# & Training set\# & Training triple\# & Validation set\# & Validation triple\# & Test set\# \\
		\hline
		Travel 			& 53,261 & 35,504 & 355,040 & 8,876 & 88,760 & 8,881 \\
		Relationships 	& 57,576 & 38,384 & 383,840 & 9,596 & 95,960 & 9,596\\
		Finance 		& 66,129 & 44,084 & 440,840 & 11,021 & 110,210 & 11,024 \\
		\hline
	\end{tabularx}
	\label{tb4}
\end{table*}

\subsection{Experimental Settings}
\subsubsection{Evaluation Metrics}
We use the same evaluation method employed by \cite{LuLi-51,ShenRong-155} to evaluate the accuracy of matching questions and answers. A set of candidate answers is created with size 6 (one positive + five negative) for each question in the test data. We compare the performance of our approach in ranking quality of the six candidate answers against that of others baselines. Discounted cumulative gain (DCG) \cite{RvelinInen-71} is employed to evaluate the ranking quality. The premise of DCG is that highly relevant documents appearing lower in a ranking list should be penalised as the graded relevance value is reduced logarithmically proportional to the position of the result. DCG accumulated at a particular rank position $p$ is defined as:
\begin{equation}
\mathrm{DCG@p} = rel_{1} + \sum_{i=2}^{p} \frac{rel_{i}}{\log_{2}(i)}
\end{equation}
where the best answer $rel = 1$, for other answers $rel = 0$. We choose DCG@1 to evaluate the precision of choosing the best answer and DCG@6 to evaluate the quality of ranking.

\subsubsection{Baseline}
We compare WEC model against Translation model (TM) \cite{JeonCroft-67}, Translation based language model (TRLM) \cite{XueJeon-87}, Okapi model \cite{JeonCroft-67} and Language model (LM) \cite{JeonCroft-67}. TM and TRLM use translation probabilities to overcome the lexical gap, while Okapi and LM only consider words that exist in both question and answer.

Given a question $q$ and answer $a$, TM \cite{JeonCroft-67} can be define as:
\begin{equation}
\begin{split}
S_{q,a} = \prod_{t\in q} {((1-\lambda) \sum_{w\in a}} & {P(t|w)P_{ml}(w|a)} \\ & +\lambda P_{ml}(t|Coll))
\end{split}
\end{equation}\textbf{}
TRLM \cite{XueJeon-87} can be define as:
\begin{equation}
	\begin{split}
		S_{q,a} = & \prod_{t\in q} {((1-\lambda)(\beta \sum_{w\in a} {P(t|w)P_{ml}(w|a)}} \\ & + (1-\beta)P_{ml}(t|a)+\lambda P_{ml}(t|Coll)))
	\end{split}
\end{equation}
where $P(t|w)$ denotes the probability that question word $t$ is the translation of answer word $w$. IBM translation model 1 is used to learn $P(t|w)$ with answer as the source and question as the target. In \cite{ShenRong-155}, word vector cosine similarity is used as word translation probabilities in TM and TRLM. Their results are also included in the experimental study.

\subsubsection{Hyperparameter}
To train the skip-gram model, we use the hyper-parameters recommended in \cite{MikolovChen-63}: the dimension of word embedding is set to 500, and the window size is 10.

CNN model contains two convolution layers labelled as C1 and C2, each convolution layer is followed by a max-pooling layers P1 and P2, and fully connected layer F. Each unit in a convolution layer is connected to a $5 \times 5$ neighbourhood in the input. Each unit in max-pooling layer is connected to a $2 \times 2$ neighbourhood in the corresponding feature map in C1. Layer C1 and M1 contains 20 feature maps each. Layer C2 and M2 contains 50 feature maps each. The fully connected layer contains 500 units and is fully connected to M2.

\subsection{Experiment Results}
\begin{table*}[!t]
	\center
	\caption{Performance of different approaches on Yahoo! Answers dataset (WEC denote sentence-level WEC function)}
	\begin{tabularx}{140mm}{|X|c|c|c|c|c|c|}
		\hline
		\multirow{2}{*}{Approach} & \multicolumn{2}{c|}{Travel} & \multicolumn{2}{c|}{Relationships} & \multicolumn{2}{c|}{Finance} \\
		\cline{2-7}
		& DCG@1 & DCG@6 & DCG@1 & DCG@6 & DCG@1 & DCG@6 \\
		\hline
		WEC + CNN & \textbf{0.761} & \textbf{0.946} & \textbf{0.709} & \textbf{0.938} & \textbf{0.780} & \textbf{0.952} \\

		WEC & 0.734 & \textbf{0.946} & 0.698 & 0.936 & 0.761 & 0.949 \\
		\hline
		TRLM &	0.727 &	0.922 & 0.683 & 0.910 & 0.755 & 0.927 \\

		TM &	0.698 &	0.914 & 0.676 & 0.912 & 0.742 & 0.926 \\
		\hline
		Okapi &	0.631 &	0.875 & 0.517 & 0.823 & 0.646 & 0.866 \\

		LM &	0.592 &	0.848 & 0.525 & 0.825 & 0.595 & 0.838 \\
		\hline
	\end{tabularx}
	\label{tb1}
\end{table*}

Table \ref{tb1} shows the Q\&A matching performance of WEC based methods, translation probability based methods and traditional retrieval methods on the Yahoo! Answer dataset. For top candidate answer precision, WEC slightly outperforms the translation probability based models. For candidate answer ranking qualities, WEC outperforms TRLM and TM. By adding CNN into the model, WEC+CNN outperforms all other models. It is possible to interpret that WEC model can perform better than TRLM and TM model, but merely using lexical level information limits its ability in selecting the best answer. Thus, WEC+CNN is able to improve the result by adding syntactical information into the model.

\begin{table*}[!t]
	\center
	\caption{Performance of different approaches on Baidu Zhidao dataset. IBM-1 denotes that translation probabilities are learned using IBM translation model 1, $cos$ denotes that translation probabilities are calculated through word vector's cosine-similarity}
	\begin{tabularx}{180mm}{|c|X|X|X|X|X|X|X|X|}
		\hline
		\multirow{3}{*}{Approach}  & \multicolumn{2}{c|}{Computers \&} 	& \multicolumn{2}{c|}{Education \&} & \multicolumn{2}{c|}{Games} 	& \multicolumn{2}{c|}{Entertainment \&}\\
		& \multicolumn{2}{c|}{Internet} 	& \multicolumn{2}{c|}{Science} & \multicolumn{2}{c|}{} 	& \multicolumn{2}{c|}{Recreation}\\
		\cline{2-9}
		& DCG@1 & DCG@6 & DCG@1 & DCG@6 & DCG@1 & DCG@6 & DCG@1 & DCG@6 \\
		\hline
		WEC+CNN & \textbf{0.826} & \textbf{0.970} & \textbf{0.870} & \textbf{0.980} & \textbf{0.703} & \textbf{0.941} & \textbf{0.780} & \textbf{0.963}	\\
		
		WEC & 0.821 & 0.968 & 0.838 & 0.975 & 0.692 & 0.937 & 0.778 & 0.962 \\
		\hline
		TRLM(IBM-1) &	0.780 &	0.937 & 0.843 & 0.948 & 0.654 & 0.894 & 0.709 & 0.918\\
		TM(IBM-1) &	0.732 &	0.925 & 0.766 & 0.931 & 0.598 & 0.876 & 0.626 & 0.892\\
		\hline
		$S$+CNN & 0.658 &	0.912 & 0.734 & 0.939 & 0.619 & 0.894 & 0.543 & 0.866\\
		\hline
		TRLM($cos$) &	0.601 &	0.885 & 0.698 & 0.924 & 0.562 & 0.865 & 0.492 & 0.843\\
		TM($cos$) &	0.596 &	0.885 & 0.691 & 0.922 & 0.560 & 0.863 & 0.486 & 0.841\\
		\hline
		Okapi &	0.567 &	0.806 & 0.702 & 0.869 & 0.467 & 0.747 & 0.446 & 0.723\\
		LM &	0.624 &	0.830 & 0.746 & 0.881 & 0.544 & 0.765 & 0.488 & 0.740\\
		\hline
	\end{tabularx}
	\label{tb2}
\end{table*}

Table \ref{tb2} shows the Q\&A matching performance of different approaches on Baidu Zhidao dataset. We find that WEC and WEC+CNN outperform all other models. Furthermore, IBM-1 based models outperform $cos$-similarity based models. This is possibly is due to the heterogeneity of question and answer, since both WEC and IBM translation model 1 can directly model the word-to-word co-occurrence probability instead of semantic similarity. On both datasets, traditional retrieval models obtain the worst result because they suffer from the lexical gap problem.

\subsection{Examples}
\begin{table*}[!t]
	\center
	\caption{Word-to-word translation examples learned from the Yahoo! Answer travel category dataset. Each column show the top 5 related answer words for a given question word. TTable denotes the type of word-to-word correlations model table used. The $cos$ denote word vector's cosine similarity.}
	\begin{tabularx}{180mm}{|c|X|X|X||X|X|X|}
		\hline
		Target &  \multicolumn{3}{c||}{where}	& \multicolumn{3}{c|}{when} \\
		\hline
		TTable & \textbf{WEC} & $\mathbf{cos}$ & \textbf{IBM model 1} & \textbf{WEC} & $\mathbf{cos}$ & \textbf{IBM model 1} \\
		\hline
		1 & middle & what & hamlets & when & before & visist \\
		\hline
		2 & southern & how & prefecture &	after & while & glacial \\
		\hline
		3 & southeastern & which & foxborough &	until & once & onward/return \\
		\hline
		4 & situated & tellme & berea 	& early & because & earthquake \\
		\hline
		5 & burundi & want & unincorporated & planting & if & feb \\
		\hline
		\hline
		Target &  \multicolumn{3}{c||}{museum}	& \multicolumn{3}{c|}{food} \\
		\hline
		TTable & \textbf{WEC} & $\mathbf{cos}$ & \textbf{IBM model 1} & \textbf{WEC} & $\mathbf{cos}$ & \textbf{IBM model 1} \\
		\hline
		1 & exhibits & galleries & rodin & vegetarian & delicious & cassoulet  \\
		\hline
		2 & musuem & monuments & montagne & seafood & cuisine & pork  \\
		\hline
		3 & planetarium & capitoline & louvre & eat & seafood & cuisine  \\
		\hline
		4 & Smithsonian & exhibits & loews 	& burgers & spicy & prolific  \\
		\hline
		5 & archaeology & musicals & chabot & cuisine & vegetarian & delft  \\
		\hline
	\end{tabularx}
	\label{tb3}
\end{table*}

To better understand the behavior of WEC, we illustrated a number of example translations from answer words to a given question word in Table \ref{tb3}. Three different methods,  e.g., WEC, $cos$-similarity, IBM Model 1, are employed to estimate the translation probabilities. Interestingly, these methods provide semantically related target words with different characters. 
To clarify this difference, consider the word ``where''. IBM model 1 provides ``hamlets'', ``prefecture'', ``foxborough'' and ``berea''. They are rarely appeared (comparing with ``middle'' and ``southern'') generic or specific name for settlement. $Cos$-similarity provides ``what'', ``how'', and ``which''. They are question words like ``where''. WEC model provides ``middle'', ``southern'', ``southeastern'', and ``situated''. These words are semantically related to the target word, and likely to appear in a suitable answer. The difference between three models reflect differences in their learning processes. 

IBM translation model 1 leverage the co-occurrence of words in parallel corpus to learn translation probabilities. The sum of translation probabilities for a question word equal to 1. Therefore, answer words with low document frequency get relatively higher translation probabilities with certain question words, because these words co-occur with a smell set of question words, hence its translation probabilities concentrate on this set of words. Skip-gram model learn embedding of words into an $n$-dimensional vector and to use these vectors to predict how likely a word is given its context. Thus, the $cos$-similarity captures the probability that a pair of words appear with similar contexts.

\begin{figure*}[t!]
	\centering
	\includegraphics[width=2\columnwidth]{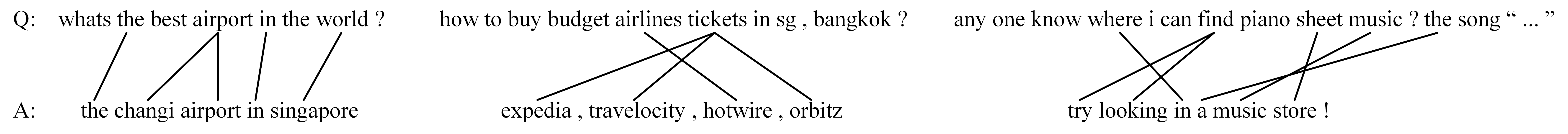}
	\caption{Real Q\&A pairs from Yahoo! Answers travel category. Relevant word pairs are link with solid line.}
	\label{picture4}
\end{figure*}

WEC tries to combines advantages of IBM model 1 and word embedding. The word vector capture distributional syntactic and semantic information via the word co-occurrence statistics \cite{BengioDucharme-163,MikolovChen-63}. Word-to-word correlation score are learned via maximizing the result of sentence-level correlation function Eq. (\ref{sentence-level function}). Meanwhile, WEC do not normalize the correlation score, which is more feasible for QA tasks.

The sentence-level correlation function Eq. (\ref{sentence-level function}) is also capable of identifying important relevance between questions and answers. As shown in the Fig. \ref{picture4}, for each word in an answer, the max operator in Eq. (\ref{sentence-level function}) chooses the most relevant word in question, based on the correlation score calculated by Eq. (\ref{word-level function}). Interestingly, both words ``try'' and ``looking'' are linked with ``find'', while the relevance between ``try'' and ``find'' is more obscure than the obvious relevance between ``looking'' and ``find''. Although, the link between ``a'' and ``the'' is inappropriate in this context, but in many other contexts, this relationship may be correct. The relation between words given a certain context is left for future work.

\section{Related Work}
\subsection{Lexical Gap Problem in CQA}
To fully use Q\&A archives in CQA systems, there are two important tasks for a newly submitted question, including question retrieval, which focuses on matching new questions with archived questions in CQA systems \cite{JeonCroft-67,XueJeon-87,CaoCong-52,CaiZhou-160,ZhouHe-156} and answer locating, which focuses finding a potentially suitable answer within a collection of candidate answers \cite{BergerCaruana-162,SurdeanuCiaramita-77,LuLi-51,ShenRong-155}. One major challenge in both tasks is the lexical gap (chasm) problem \cite{DBLP:confChenJYYZ16}. Potential solutions to overcome this difficulty include 1) query expansion, 2) statistical translation, 3) latent variable models  \cite{BergerCaruana-162}. 

Much importance has been attached to statistical translation in the literature. Classical methods include translation model (TM) \cite{JeonCroft-67} and translation language model (TRLM) \cite{XueJeon-87}. Both use IBM translation model 1 to learn the translation probabilities between question and answer words. Apart from word-level translation, phrase-level translation for question and answer retrieval has also achieved promising results \cite{CaiZhou-160}. 

Latent variable models also attract much research in recent years. Proposals have been made to learn the latent topics aligned across the question-answer pairs to bridge the lexical gap, on the assumption that question and its answer should share a similar topic distribution \cite{CaiZhou-160,JiXu-161}. Furthermore, inspired by the recent success of word embedding, several approaches have been proposed to leverage the advantages of the vector representation to overcome the lexical gap \cite{ShenRong-155,ZhouHe-156}. 

Different from previous models, our work aims at combining the idea of both statistical translation and latent variable model. We proposed a latent variable model, but parameters are learned to model the word-level translation probabilities. As a result, we can keep the generalisability of latent variable model, while achieving better precision than a brutal statistical translation model and provide more reasonable results in word-to-word correlation examples.

\subsection{Translation Matrix}
Distributed representations for words have proven its success in many domain applications. Its main advantage is that the representations of similar words are close in the vector space, which makes generalisation to novel patterns easier and model estimation more robust. Successful follow-up work includes application to statistical language modelling \cite{BengioDucharme-163,MikolovChen-63}. 

Inspired by vector representation of words, the translation matrix has been proposed to map vector representation $x$ from one language space to another language space, using cosine similarity as a distance metric \cite{DBLP:journals/corr/MikolovLS13}. Our word-level WEC model uses the same translation functions to map vector $x$ from answer semantic space to question semantic space. We further propose a sentence-level WEC model to calculate the Q\&A matching probability, and a method to learn the translation matrix through maximising the matching accuracy in a parallel Q\&A corpus.

Similarly neural tensor network (NTN) is also implemented to model relational information \cite{SocherChen-169,qiu2015convolutional}. A tensor matrix is employed to seize the relationship between vectors. The NTN's main advantage is that it can relate two inputs multiplicatively instead of only implicitly through non-linearity as with standard neural networks where the entity vectors are simply concatenated \cite{SocherChen-169}. Our model is conceptually similar to NTN and use a translation matrix to model the word-to-word relation in Q\&A pairs. Similar to NTN, the translation matrix in our model makes it possible to explicitly relate the two inputs, and $cos$ in Eq. (\ref{word-level function}) adds non-linearity.

\section{Conclusion and Future Work}
This paper presents a new approach for Q\&A matching in CQA services. In order to solve the lexical gap between question and answer, a word embedding based correlation (WEC) model is proposed, where the co-occurrence relation between words in parallel text is represented as a matrix (or a set of matrices). Given a random pair of words, WEC model can score their co-occurrence probability in Q\&A pairs like the previous translation model based approach. And it also leverages the continuity and smoothness of continuous space word representation to deal with new pairs of words that are rare in the training parallel text. Our experiments show that WEC and WEC+CNN outperform state-of-the-art models.

There are several interesting directions which deserve further exploration in the future. It is possible to apply this model in question-question matching tasks, or multi-language question retrieval task. It is also interesting to explore the possibility of using this approach to solve other parallel detection problems (e.g., comment selection on a given tweet).

\section*{Acknowledgment}
This work was partially supported by the National Natural Science Foundation of China (No. 61332018), the National Department Public Benefit Research Foundation of China (No. 201510209), and the Fundamental Research Funds for the Central Universities.

\bibliographystyle{named}
\bibliography{reference}

\end{document}